\documentclass[11pt,a4paper]{article}
\usepackage[hyperref]{acl2019}
\usepackage{times}
\usepackage{latexsym}
\usepackage{multirow}
\usepackage{url}
\usepackage{graphicx}
\usepackage{bm}
\usepackage{amsmath}
\usepackage{amssymb}
\usepackage{amstext}
\usepackage{mathrsfs}
\usepackage{array}
\usepackage{booktabs}
\usepackage{url}

\aclfinalcopy 
\def\aclpaperid{461} 

\title{Multi-Level Matching and Aggregation Network \\ for Few-Shot Relation Classification}

\author{Zhi-Xiu Ye, Zhen-Hua Ling\thanks{\hspace{0.1cm} Corresponding author: Zhen-Hua Ling.} \\
National Engineering Laboratory for Speech and Language Information Processing, \\
University of Science and Technology of China \\
{\tt zxye@mail.ustc.edu.cn}, {\tt zhling@ustc.edu.cn}
}

\date{}

\begin{document}
\maketitle

\begin{abstract}
This paper presents a multi-level matching and aggregation network (MLMAN) for few-shot relation classification.
Previous studies on this topic adopt prototypical networks, which calculate the embedding vector of a query instance and
the prototype vector of each support set independently.
In contrast, our proposed MLMAN model encodes the query instance and each support set in an interactive way
by considering their matching information at both local and instance levels.
The final class prototype for each support set is obtained by attentive aggregation over the representations of its support instances, where the weights are calculated using the query instance.
Experimental results demonstrate the effectiveness of our proposed methods, which achieve a new state-of-the-art performance on the FewRel dataset\footnote{The code is available at \url{https://github.com/ZhixiuYe/MLMAN}.}.
\end{abstract}

\section{Introduction}

Relation classification (RC) is a fundamental task in natural language processing (NLP), which aims to identify the semantic relation between two entities in text.
For example, the instance ``[London]$_{e1}$ is the capital of [the UK]$_{e2}$" expresses the relation \emph{capital\_of} between the two entities \textbf{London} and \textbf{the UK}.

\begin{table}[t]
	\small
	\begin{center}
		\begin{tabular}{p{7.1cm}}
			\toprule
			\textbf{Support Set} \\
			\midrule
			class A: \emph{mother} \\
			\underline{instance \#1} The Queen Consort [Jetsun Pema]$_{e2}$ gave birth to a son on 5 February 2016 , [Jigme Namgyel Wangchuck]$_{e1}$.\\
			\underline{instance \#2} He married the American actress [Cindy Robbins]$_{e2}$ and was stepfather to her daughter , [Kimberly Beck]$_{e1}$.\\
			\underline{instance \#3} Edgar married actress [Moyna Macgill]$_{e2}$ and became the father of [Angela Lansbury]$_{e1}$.\\
			\underline{instance \#4} In 1845 , [Cemile Sultan]$_{e1}$ 's mother , Empress [Düzdidil Kadın]$_{e2}$, died.\\
			\underline{instance \#5} Bo 's wife [Gu Kailai]$_{e2}$ traveled with their son [Bo Guagua]$_{e1}$ to Britain.\\
			\midrule
			class B: \emph{member\_of} ...\\
			\midrule
			class C: \emph{father} ...\\
			\midrule
			class D: \emph{sport} ...\\
			\midrule
			class E: \emph{voice\_type} ...\\
			\bottomrule
			\toprule
			\textbf{Query Instance} \\
			\midrule
			He was married to [Eva Funck]$_{e2}$ and they have a son [Gustav]$_{e1}$ .\\
			\bottomrule
		\end{tabular}
	\end{center}
	\caption{A data example of 5-way-5-shot relation classification in FewRel development set. The correct relation class for the query instance is class A: \emph{mother}. The instances for other relation classes are omitted for saving space.}
	\label{example}
\end{table}

Some conventional relation classification methods \citep{Bethard2007Temporal,Zelenko2002Kernel} adopted supervised training and suffered from the lack of large-scale manually labeled data.
To address this issue, the distant supervision method \citep{mintz2009distant} was proposed which annotated training data by heuristically aligning knowledge bases (KBs) and texts.
However, the long-tail problem in KBs \citep{Xiong2018One,Han2018FewRel} still exists and makes it hard to classify the relations with very few training samples.

This paper focuses on the few-shot relation classification task, which was designed to address the long-tail problem.
In this task, only few (e.g., 1 or 5) support instances are given for each relation, as  shown by an example in Table \ref{example}.

The few-shot learning problem has been studied extensively in computer vision (CV) field.
Some methods adopt meta-learning architectures \citep{santoro2016meta,ravi2016optimization,finn2017model,munkhdalai2017meta}, which learn fast-learning abilities from previous experiences (e.g., training set) and then rapidly generalize to new concepts (e.g., test set).
Some other methods use metric learning based networks \citep{koch2015siamese,vinyals2016matching,snell2017prototypical}, which learn the distance distributions among classes.
A simple and effective metric-based few-shot learning method is prototypical network \citep{snell2017prototypical}.
In a prototype network, query and support instances are encoded into an embedding space independently.
Then,  a prototype vector for each class candidate is derived as the mean of its support instances in the embedding space.
Finally, classification is  performed by calculating the distances between the embedding vector of the query and all class prototypes.
This prototype network method has also been applied to few-shot relation classification recently \citep{Han2018FewRel}.

This paper proposes a multi-level matching and aggregation network (MLMAN) for few-shot relation classification.
Different from prototypical networks, which represent support sets without dependency on query instances,
our proposed MLMAN model encodes each query instance and each support set in an interactive way
by considering their matching information at both local and instance levels.
At local level, the local context representations of a query instance and a support set are softly matched toward each other
following the sentence matching framework \citep{Chen2017Enhanced}.
Then, the matched local representations are aggregated into an embedding vector for each query and each support instance using max and average pooling.
At instance level, the matching degree between the query instance and each of the support instances is calculated via a multi-layer perceptron (MLP).
Taking the matching degrees as weights, the instances in a support set are aggregated to form the class prototype for final classification.
All these matching and aggregation layers in the MLMAN model are estimated jointly using training data.
Since the representations of the support instances in each class are expected to be close with each other,
an auxiliary loss function is further designed to measure the inconsistency among all support representations in each class.

In summary, our contributions in this paper are three-fold.
First, a multi-level matching and aggregation network is proposed to encode query instances and class prototypes in an interactive fashion.
Second, an auxiliary loss function measuring the consistency among support instances is designed.
Third, our method achieves a new state-of-the-art performance on FewRel, a public few-shot relation classification dataset.

\section{Related Work}

\subsection{Relation Classification}

Relation classification is to identify the semantic relation between two entities in one sentence.
In recently years, neural networks have been widely applied to deal with this task.
\citet{Zeng2014Relation} employed  position features and convolutional neural networks (CNNs) to capture the structure and contextual information respectively.
Then, a max pooling operation was adopted to determine the most useful features.
\citet{Wang2016Relation} proposed multi-level attention CNNs, which captured both entity-specific attention and relation-specific pooling attention
 in order to better discern patterns in heterogeneous contexts.
\citet{Zhou2016Attention} proposed attention-based bidirectional long short-term memory networks (AttBLSTMs) to capture the most important semantic information in a sentence.
All of these methods require a large amount of training data and can't quickly adapt to a new class that has never been seen.

\subsection{Metric Based Few-Shot Learning}

In few-shot learning paradigm, a classifier is required to generalize to new classes with only a small number of training samples.
The metric based approach aims to learn a set of projection functions that take support and query samples from the target problem and classify them in a feed forward manner.
This approach has lower complexity and is easier for implementation than meta-learner based approach \citep{ravi2016optimization,finn2017model,santoro2016meta,munkhdalai2017meta}.

Some metric based few-shot learning methods have been developed for computer vision (CV)  tasks,
and all these methods encoded each support or query image to a vector independently for classification.
\citet{koch2015siamese} proposed a method for learning siamese neural networks, which employed an unique structure to encode both support and query samples respectively and one more layer computing the induced distance metric between the pair.
\citet{vinyals2016matching} proposed to learn a matching network augmented with attention and external memories. And also, an episode-based training procedure was proposed, which was based on a principle that test and training conditions must match and has been adopted by many following studies.
\citet{snell2017prototypical} proposed prototypical networks that learn a metric space in which classification can be performed by computing distances to prototype representations of all classes, and the prototype representation of each class was the mean of all its support samples.
\citet{garcia2017few} defined a graph neural network architecture to assimilate generic message-passing inference algorithms, which generalized above three models.

Regarding with few-shot relation classification, \citet{Han2018FewRel} adopted prototypical networks to build baseline models on the FewRel dataset.
\citet{gao2019hybrid} proposed hybrid attention-based prototypical networks to handle noisy training samples in few-shot learning.
In this paper, we improve the conventional prototypical networks for few-shot relation classification by encoding the query instance and class prototype interactively through multi-level matching and aggregation.

\subsection{Sentence Matching}

Sentence matching is essential for many NLP tasks, such as natural language inference (NLI) \citep{Bowman2015A} and response selection \citep{Lowe2015The}.
Some sentence matching methods mainly rely on sentence encoding \citep{mueller2016siamese,Conneau2017Supervised,Chen2018Enhancing}, which encode a pair sentences independently and then transmit their embeddings into a classifier, such as a neural network, to decide the relationship between them.
Some other methods are based on joint models \citep{Chen2017Enhanced,gong2017natural,kim2018semantic},
which use cross-features to represent the local (i.e., word-level and phrase-level) alignments for better performance.
In this paper, we follow the joint models to achieve the local matching between a query instance and the support set for a class.
The difference between our task and the other sentence matching tasks mentioned above is that, our goal is to match a sentence to a set of sentences,
instead of to another sentence \citep{Bowman2015A} or to a sequence of sentences \citep{Lowe2015The}.

\section{Task Definition}
In few-shot relation classification, we are given two datasets, $\mathscr{D}_{meta-train}$ and $\mathscr{D}_{meta-test}$.
Each dataset consists of a set of samples $( x, p, r )$, where $x$ is a sentence composed of $T$ words and the $t$-th word is $w_t$, $p = (p_1,p_2)$ indicate the positions of two entities, and $r$ is the relation label of the instance $(x, p)$.
These two datasets have their own relation label spaces that are disjoint with each other.
Under few-shot configuration,  $\mathscr{D}_{meta-test}$ is splited into two parts, $\mathscr{D}_{test-support}$ and $\mathscr{D}_{test-query}$.
If $\mathscr{D}_{test-support}$ contains $K$ labeled samples for each of $N$ relation classes, this target few-shot problem is named $N$-way-$K$-shot.
$\mathscr{D}_{test-query}$ contains test samples, each labeled with one of the $N$ classes.
Assuming that we  only have $\mathscr{D}_{test-support}$ and $\mathscr{D}_{test-query}$, we can train a model using $\mathscr{D}_{test-support}$ and evaluate its performance on $\mathscr{D}_{test-query}$.
But limited by the number of support samples (i.e,., $N \times K$), it is hard to train a good model from scratch.

Although $\mathscr{D}_{meta-train}$ and $\mathscr{D}_{meta-test}$ have disjoint relation label spaces,
$\mathscr{D}_{meta-train}$ can also been utilized to help the few-shot relation classification on  $\mathscr{D}_{meta-test}$.
One approach is the paradigm proposed by \citet{vinyals2016matching}, which obey an important machine learning principle that test and train conditions must match.
That's to say, we also split $\mathscr{D}_{meta-train}$ into two parts, $\mathscr{D}_{train-support}$ and $\mathscr{D}_{train-query}$, and mimic the few-shot learning settings at training stage.
In each training iteration, $N$ classes are randomly selected from $\mathscr{D}_{train-support}$, and $K$ support instances are randomly selected  from each class.
In this way, we construct the train-support set $S = \{s_k^i; i=1,...,N,k=1,...,K\}$, where $s_k^i$ is the k-th instance in class $i$.
And also, we randomly select $R$ samples from the remaining samples of those $N$ classes and construct the train-query set $Q = \{(q_j, l_j); j=1,...,R\}$, where $l_j \in \{1, ..., N\}$ is the label of instance $q_j$.

\begin{figure*}[!t]
	\centering
	\includegraphics[width=6.2in]{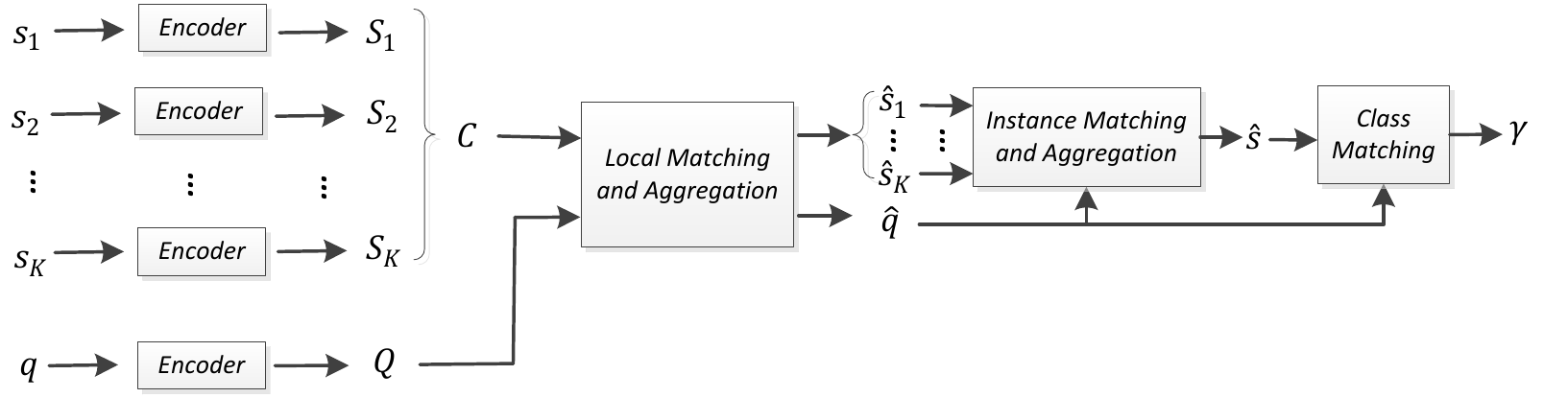}
	\caption{
		The framework of our proposed MLMAN model.
	}
	\label{framework}
\end{figure*}

Just like conventional prototypical networks, we expect to minimize the following objective function at training time
\begin{equation}
	{\rm J}_{match} = - \frac{1}{R}\sum_{(q,l)\in Q} {\rm P}(l|S,q),
\label{obj}
\end{equation}
and ${\rm P}(l|S, q)$ is defined as
\begin{equation}
	{\rm P}(l|S, q) = \frac{{\rm exp}(f(\{s_k^{l}\}_{k=1}^{K}, q))}{\sum_{i=1}^{N}{\rm exp}(f(\{s_{k}^{i}\}_{k=1}^{K}, q))}.
	\label{functionf}
\end{equation}
The function $f(\{s_k^{i}\}_{k=1}^{K}, q)$ is to calculate the matching degree between the query instance $q$ and the set of support instances $\{s_k^{i}\}_{k=1}^{K}$.
How to design this function is the focus of this paper.

\section{Methodology}
In this section, we will introduce our proposed multi-level matching and aggregation network (MLMAN) for modeling $f(\{s_k^i\}_{k=1}^{K}, q)$.
For simplicity, we will discard the superscript $i$ of $s_{k}^{i}$ from Section 4.1 to Section 4.4.
The framework of our proposed MLMAN model  is shown in Fig. \ref{framework}, which has four main modules.

\begin{itemize}
	\item \textbf{Context Encoder.} Given a sentence and the positions of two entities within this sentence, CNNs \citep{Zeng2014Relation} are adopted to derive the local context representations of each word in the sentence.
	
	\item \textbf{Local Matching and Aggregation.} 
	Similar to \citep{Chen2017Enhanced}, given the local representation of a query instance and the local representations of $K$ support instances, the attention method is employed to collect local matching information between them.
	Then, the matched local representations are aggregated to represent each instance as an embedding vector.
	
	\item \textbf{Instance Matching and Aggregation.} 
	The matching information between a query instance and each of the $K$ support instances are calculated using an MLP.
    Then, we take the matching degrees as weights to sum the representations of support instances in order to get the class prototype.
	
	\item \textbf{Class Matching.} 
    An MLP is built to calculate the matching score between the representations of the query instance and the class prototype.
\end{itemize}

More details of these four modules will be introduced in the following subsections.

\subsection{Context Encoder}

For a query or support instance, each word $w_t$ in the sentence $x$ is first mapped into a $d_w$-dimensional word embedding $\mathbf{e}_{t}$ \citep{Pennington2014Glove}.
In order to describe the position information of the two entities in this instance, the position features (PFs) proposed by \citet{Zeng2014Relation} are also adopted in our work.
Here, PFs describe the relative distances between current word and the two entities, and are further mapped into two vectors $\mathbf{p}_{1t}$ and $\mathbf{p}_{2t}$ of $d_p$ dimensions.
Finally, these three vectors are concatenated to get the word representation $\mathbf{w}_{t} = [\mathbf{e}_{t}; \mathbf{p}_{1t}; \mathbf{p}_{2t}]$ of $d_{w} + 2d_{p}$ dimensions, and the instance can be written as $\mathbf{W} \in \mathbb{R}^{T \times (d_{w} + 2d_{p})}$.

The most popular models for local context encoding are recurrent neural networks (RNNs) with  long short-term memories (LSTMs) \citep{Hochreiter1997Long} and convolutional neural networks (CNNs) \citep{Kim2014Convolutional}.
In this paper, we employ CNNs to build the context encoder.
For an input instance $\mathbf{W} \in \mathbb{R}^{T \times (d_{w} + 2d_{p})}$, we input it into a CNN with $d_c$ filters.
The output from the CNN is a matrix with $T \times d_c$ dimensions.
In this way, the context representations of the query instance $\mathbf{Q} \in \mathbb{R}^{T_q \times d_c}$ and
the context representations of support instances $\{\mathbf{S}_{k} \in \mathbb{R}^{T_{k} \times d_c}; k=1,...,K\}$ are obtained,
where $T_q$ and $T_k$ are the sentence lengths of the query sentence and the $k$-th support sentence respectively.

\subsection{Local Matching and Aggregation}

In order to get the matching information between $\mathbf{Q}$ and $\{\mathbf{S}_{k}; k=1,...,K\}$,
we first concatenate the $K$ support instance representations into one matrix as follow
\begin{equation}
	\mathbf{C} = {\rm concat} (\{\mathbf{S}_{k}\}_{k=1}^K),
	\label{concat}
\end{equation}
where $\mathbf{C} \in \mathbb{R}^{T_s \times d_c }$ with $T_s = \sum_{k=1}^{K}{T_{k}}$.
Then, we collect the matching information between $\mathbf{Q}$ and $\mathbf{C}$ and calculate their matched representations $\widetilde{\mathbf{Q}}$ and $\widetilde{\mathbf{S}}$ as follows
\begin{equation}
	\alpha_{mn} = \mathbf{q}_m^{\top}\mathbf{c}_n,
\end{equation}
\begin{equation}
	\widetilde{\mathbf{q}}_m = \sum_{n=1}^{T_s} \frac {{\rm exp}(\alpha_{mn})} {\sum_{n'=1}^{T_s} {\rm exp} (\alpha_{mn'})} \mathbf{c}_n,
	\label{m}
\end{equation}
\begin{equation}
	\widetilde{\mathbf{c}}_n = \sum_{m=1}^{T_q} \frac {{\rm exp}(\alpha_{mn})} {\sum_{m'=1}^{T_q} {\rm exp} (\alpha_{m'n})} \mathbf{q}_m,
	\label{n}
\end{equation}
where $m \in \{ 1,..., T_q\}$ in Eq. (\ref{m}), $n \in \{ 1,...,T_s \}$ in Eq. (\ref{n}),
${\mathbf{q}}_m$ and $\widetilde{\mathbf{q}}_m$ are the $m$-th rows of ${\mathbf{Q}}$ and $\widetilde{\mathbf{Q}}$ respectively,
and ${\mathbf{c}}_n$ and $\widetilde{\mathbf{c}}_n$ are the $n$-th rows of ${\mathbf{C}}$ and $\widetilde{\mathbf{C}}$ respectively.

Next, the original representations and the matched representations are fused utilizing a ReLU layer as follows,
\begin{align}
	\bar{\mathbf{Q}} &= {\rm ReLU} ([\mathbf{Q}; \widetilde{\mathbf{Q}}; |\mathbf{Q}-\widetilde{\mathbf{Q}}|; \mathbf{Q} \odot \widetilde{\mathbf{Q}}]\mathbf{W}_1),\\
	\bar{\mathbf{C}} &= {\rm ReLU} ([\mathbf{C}; \widetilde{\mathbf{C}}; |\mathbf{C}-\widetilde{\mathbf{C}}|; \mathbf{C} \odot \widetilde{\mathbf{C}}]\mathbf{W}_1),
\end{align}
where $\odot$ is the element-wise product and $\mathbf{W}_1 \in \mathbb{R}^{4d_c \times d_h}$ is the weight matrix at this layer for reducing dimensionality.
$\bar{\mathbf{C}}$ is further split into $K$ representations $\{\bar{\mathbf{S}}_k\}_{k=1}^K$ corresponding to the $K$ support instances where $\bar{\mathbf{S}}_k \in \mathbb{R}^{T_{k} \times d_h}$.
All $\bar{\mathbf{S}}_k$ and $\bar{\mathbf{Q}}$ are fed into a single-layer Bi-directional LSTM (BLSTM) with $d_h$ hidden units along each direction to obtain the final local matching results $\widehat{\mathbf{S}}_k \in \mathbb{R}^{T_{k} \times 2 d_h}$ and $\widehat{\mathbf{Q}}\in \mathbb{R}^{T_{q} \times 2 d_h}$.

Local aggregation aims to convert the results of local matching into a single vector for each query and each support instance.
In this paper, we employ a max pooling together with an average pooling, and concatenate their results into one vector$\widehat{\mathbf{s}}_k$ or $\widehat{\mathbf{q}}$.
The calculations are as follows,
\begin{align}
		\widehat{\mathbf{s}}_k =& [{\rm max}(\widehat{\mathbf{S}}_k); {\rm ave}(\widehat{\mathbf{S}}_k)], \forall  k \in \{1,...,K\},\\
	    \widehat{\mathbf{q}}   =& [{\rm max}(\widehat{\mathbf{Q}}); {\rm ave}(\widehat{\mathbf{Q}})],
\end{align}
where $\{\widehat{\mathbf{s}}_k , \widehat{\mathbf{q}}\} \in \mathbb{R}^{4d_h}$.

\subsection{Instance Matching and Aggregation}
Similar to conventional prototypical networks \cite{snell2017prototypical}, our proposed method calculates class prototype $\widehat{\mathbf{s}}$ via the representations of all support instances  in this class, i.e., $\{ \widehat{\mathbf{s}}_k \}_{k=1}^K$.
However, instead of using a naive mean operation, we aggregate instance-level representations via attention over $\{ \widehat{\mathbf{s}}_k \}_{k=1}^K$,
where each weight is derived from the instance matching score between $\widehat{\mathbf{s}}_k$ and $\widehat{\mathbf{q}}$.
The matching function is as follow,
\begin{equation}
	\beta_k = \mathbf{v}^{\top}({\rm ReLU}(\mathbf{W}_2[\widehat{\mathbf{s}}_k; \widehat{\mathbf{q}}])),
	\label{matching}
\end{equation}
where $\mathbf{W}_2 \in \mathbb{R}^{d_h \times 8d_h}$ and $\mathbf{v} \in \mathbb{R}^{d_h}$.
$\beta_k$ describes the instance-level matching degree between the query instance $q$ and the support instance $s_k$.
Then, all $\{\widehat{\mathbf{s}}_k\}_{k=1}^K$ are aggregated into one vector $\widehat{\mathbf{s}}$ as
\begin{equation}
	\widehat{\mathbf{s}} = \sum_{k=1}^{K}\frac{{\rm exp}(\beta_k)}{\sum_{k'=1}^{K}{\rm exp}(\beta_k')} \widehat{\mathbf{s}}_k,
	\label{supportatt}
\end{equation}
and $\widehat{\mathbf{s}}$ is the class prototype.

\subsection{Class Matching}
After the class prototype $\widehat{\mathbf{s}}$ and the embedding vector of the query instance $\widehat{\mathbf{q}}$ have been determined, the class-level matching function $f(\{ s_k \}_{k=1}^K, q)$ in Eq. (2) is defined as
\begin{equation}
	f(\{ s_k \}_{k=1}^K, q) = \mathbf{v}^{\top}({\rm ReLU}(\mathbf{W}_2[\widehat{\mathbf{s}}; \widehat{\mathbf{q}}])).
	\label{matching2}
\end{equation}
Eq. (11) and (13) have the same form.
In our experiments, sharing the weights $\mathbf{W}_2$ and $\mathbf{v}$ in these two equations, i.e., employing the exactly same function for both instance-level and class-level matching in each training iteration, lead to better performance.

\subsection{Joint Training with Inconsistency Measurement}
If the representations of all support instances in a class are far away from each other, it could become difficult for the derived class prototype to capture the common characteristics of all support instances.
Therefore, a function which measures the inconsistency among the set of support instances is designed.
In order to avoid the high complexity of directly comparing every two support instances in a class, we calculate the inconsistency measurement as the average Euclidean distance
between the support instances and the class prototype as
\begin{equation}
{\rm J}_{incon} = \frac{1}{NK} \sum_{i=1}^{N}\sum_{k=1}^{K}||\widehat{\mathbf{s}}^i_k - \widehat{\mathbf{s}}^i||_2^2,
\label{proto}
\end{equation}
where $i$ is the  class index and $||\cdot||_2$ calculates the 2-norm of a vector.

By combining Eqs. (\ref{obj}) and (\ref{proto}), the final objective function for training the whole model is defined as
\begin{equation}
	{\rm J} = {\rm J}_{match} + \lambda {\rm J}_{incon},
\end{equation}
where $\lambda$ is a hyper-parameter and was set as 1 in our experiments without any tuning.

\begin{table*}[t!]
	\begin{center}
		\small
		\begin{tabular}{p{2.0in}|cccc}
			\toprule
			\bf Model                                  & 5 Way 1 Shot     & 5 Way 5 Shot     & 10 Way 1 Shot    & 10 Way 5 Shot    \\
			\midrule
			Meta Network \cite{Han2018FewRel}          & $64.46 \pm 0.54$ & $80.57 \pm 0.48$ & $53.96 \pm 0.56$ & $69.23 \pm 0.52$ \\
			GNN \cite{Han2018FewRel}                  & $66.23 \pm 0.75$ & $81.28 \pm 0.62$ & $46.27 \pm 0.80$ & $64.02 \pm 0.77$ \\
			SNAIL \cite{Han2018FewRel}                 & $67.29 \pm 0.26$ & $79.40 \pm 0.22$ & $53.28 \pm 0.27$ & $68.33 \pm 0.25$ \\
			Prototypical Network \cite{Han2018FewRel}  & $69.20 \pm 0.20$ & $84.79 \pm 0.16$ & $56.44 \pm 0.22$ & $75.55 \pm 0.19$ \\
			Proto-HATT \cite{gao2019hybrid}            & - -              & $90.12 \pm 0.04$ & - -              & $83.05 \pm 0.05$ \\
			\midrule
			\midrule
			MLMAN                                      & $82.98 \pm 0.20$ & $92.66 \pm 0.09$ & $75.59 \pm 0.27$ & $87.29 \pm 0.15$ \\
			\bottomrule
		\end{tabular}
	\end{center}
	\caption{ Accuracies (\%) of different models on FewRel test set. }
	\label{test}
\end{table*}

\section{Experiments}
\subsection{Dataset and Evaluation Metrics}
The few-shot relation classification dataset FewRel\footnote{\url{https://thunlp.github.io/fewrel.html}.} was adopted in our experiments.
This dataset was first generated by distant supervision and then filtered by crowdsourcing to remove noisy annotations.
The final FewRel dataset consists of 100 relations, each has 700 instances.
The average number of tokens in each sentence is 24.99, and there are 124,577 unique tokens in total.
The 100 relations are split into 64, 16 and 20 for training, validation and test respectively.

Our experiments investigated four few-shot learning configurations, 5 way 1 shot, 5 way 5 shot, 10 way 1 shot, and 10 way 5 shot, which were the same as \citet{Han2018FewRel}.
According to the official evaluation scripts\footnote{\url{https://thunlp.github.io/fewrel.html}.}, 
all results given by our experiments were the mean and standard deviation values of 10 training repetitions, and were tested using 20,000 independent samples.

\subsection{Training Details and Hyperparameters}
All of the hyperparameters used in our experiments are listed in Table \ref{hyper}.
The 50-dimensional Glove word embeddings released by \citet{Pennington2014Glove} \footnote{\url{https://nlp.stanford.edu/projects/glove/}.}  were adopted in the context encoder and were fixed during training.
For the unknown words, we just replaced them with an unique special token $<$UNK$>$ and fixed its embedding as a zero vector.
Previous study \citep{munkhdalai2017meta} found that the models trained on harder tasks may achieve better performances than using the same
configurations at both training and test stages.
Therefore, we set $N = 20$ to construct the train-support sets for 5-way and 10-way tasks.
In our experiments, grid searches among $d_c \in \{100,150,200,250\}$, $d_h \in \{100,150,200,250\}$ and $R \in \{5,10,15\}$ were conducted to determine their optimal values.
For optimization, we employed mini-batch stochastic gradient descent (SGD) with the initial learning rate of 0.1.
The learning rate was decayed to one tenth every 20,000 steps.
And also, dropout layers \cite{hinton2012improving} were inserted before CNN and LSTM layers and the drop rate was set as 0.2.

\begin{table}[t!]
	\small
	\begin{center}
		\begin{tabular}{p{1.1in}<{\centering}|p{1.1in}<{\centering}|p{0.3in}<{\centering}}
			\hline
			\bf Component                            & \bf Parameter           & \bf Value   \\ \hline
			word embedding                           & dimension               & 50          \\ \hline
			{\multirow{2}{*}{position feature}}      & max relative distance   & $\pm$40     \\ \cline{2-3}
			& dimension               & 5           \\ \hline
			{\multirow{2}{*}{CNN}}                   & window size             & 3           \\ \cline{2-3}
			& filter number $d_c$     & 200         \\ \hline
			dropout                                  & dropout rate            & 0.2         \\ \hline
			unidirectional LSTM                      & hidden size $d_h$       & 100         \\ \hline
			{\multirow{5}{*}{optimization}}          & strategy                & SGD         \\ \cline{2-3}
			& learning rate           & 0.1         \\ \cline{2-3}
			& size of query set $R$   & 5           \\ \cline{2-3}
			& $N_{train}$             & 20          \\ \cline{2-3}
			& $\lambda$               & 1           \\
			\hline
		\end{tabular}
	\end{center}
	\caption{\label{hyper}Hyper-parameters of the models built in our experiments.}
\end{table}

\begin{table*}[t!]
	\begin{center}
		\small
		\begin{tabular}{p{1.6in}|c|cccc}
			\toprule
			\bf Model                               &No.   & 5 Way 1 Shot      & 5 Way 5 Shot     & 10 Way 1 Shot    & 10 Way 5 Shot    \\
			\midrule
			MLMAN                                   &1     & $79.01 \pm 0.20$  & $88.86 \pm 0.20$ & $67.37 \pm 0.19$ & $80.07 \pm 0.18$ \\
			\midrule
			\quad -${\rm J}_{incon}$                &2     & $79.01 \pm 0.20$  & $88.33 \pm 0.15$ & $67.37 \pm 0.19$ & $79.38 \pm 0.22$ \\
			\quad IM(shared $\to$ untied)           &3     & $79.01 \pm 0.20$  & $86.77 \pm 0.19$ & $67.37 \pm 0.19$ & $77.66 \pm 0.09$ \\
			\quad IA(att. $\to$ max.)               &4     & $79.01 \pm 0.20$  & $87.84 \pm 0.13$ & $67.37 \pm 0.19$ & $78.86 \pm 0.15$ \\
			\quad IA(att. $\to$ ave.)               &5     & $79.01 \pm 0.20$  & $87.48 \pm 0.17$ & $67.37 \pm 0.19$ & $78.58 \pm 0.23$ \\
			\quad\quad -${\rm J}_{incon}$           &6     & $79.01 \pm 0.20$  & $86.23 \pm 0.22$ & $67.37 \pm 0.19$ & $77.36 \pm 0.26$ \\
			\quad\quad\quad LM(-concatenation)      &7     & $79.01 \pm 0.20$  & $85.48 \pm 0.28$ & $67.37 \pm 0.19$ & $74.56 \pm 0.36$ \\
			\quad\quad\quad CM(MLP $\to$ ED)        &8     & $76.52 \pm 0.23$  & $81.91 \pm 0.13$ & $62.89 \pm 0.13$ & $69.41 \pm 0.15$ \\
			\quad\quad\quad -LM                     &9     & $74.13 \pm 0.16$  & $82.73 \pm 0.16$ & $59.71 \pm 0.22$ & $70.23 \pm 0.23$ \\
			\quad\quad\quad\quad CM(MLP $\to$ ED)   &10    & $75.42 \pm 0.23$  & $82.36 \pm 0.07$ & $62.54 \pm 0.26$ & $70.45 \pm 0.11$ \\
			\bottomrule
		\end{tabular}
	\end{center}
	\caption{ Accuracies (\%) of different models on FewRel development set. Here, IM stands for instance matching, IA stands for instance aggregation, LM stands for the local matching, CM stands for the class matching, MLP stands for  multi-layer perceptrons and ED stands for Euclidean distance.}
	\label{ablation}
\end{table*}

\subsection{Comparison with Previous Work}
Table \ref{test} shows the results of different models tested on FewRel test set.
The results of the first four models, Meta Network \cite{munkhdalai2017meta}, GNN \cite{garcia2017few}, SNAIL \cite{Mishra2018A}, Prorotypical Network \cite{snell2017prototypical}, were reported by \citet{Han2018FewRel}.
These models were initially proposed for image classification.
\citet{Han2018FewRel} just replaced their image encoding module with an instance encoding module and kept other modules unchanged.
Proto-HATT \citep{gao2019hybrid} added hybrid attention mechanism to prototypical networks, mainly focusing on improving the performance on few-shot relation classification with $N>1$.
From Table \ref{test}, we can see that our proposed MLMAN model outperforms all other models by a large margin, which shows the effectiveness of considering the interactions between query instance and support set at multiple levels.

\subsection{Ablation Study}
In order to evaluate the contributions of individual model components, ablation studies were conducted.
Table \ref{ablation} shows the performance of our model and its ablations on the development set of FewRel.
Considering that the first 6 ablations only affected the few-shot learning tasks with $N>1$, model 2 to model 7 achieved exactly
the same performance as the complete model (i.e., model 1) under 5 way 1 shot and 10 way 1 shot configurations.

\subsubsection{Instance Matching and Aggregation}

First, the attention-based instance  aggregation introduced in Section 4.3 was replaced with a max pooling (model 4) or an average pooling (model 5).
We can see that the model with instance-level attentive aggregation (model 1) outperformed the ones using a max pooling (model 4) or an average pooling (model 5) on 5-shot tasks.
 Their difference were significantly at 1\% significance level in t-test.
 The advantage of attentive pooling is that the weights of integrating all support instances can be determined dynamically according to the query.
 For example, when conducting instance matching and aggregation between the query instance and the support set in Table \ref{example},
 the weights of the 5 instances in class A were 0.03, 0.46, 0.25, 0.08 and 0.18 respectively.
 Instance \#2 achieved the highest weight because it had the best similarity with the query instance and was considered as the most helpful one
 when matching the query instance with class A.

Then, the effectiveness of sharing the weight parameters in Eqs. (\ref{matching}) and (\ref{matching2}) was evaluated by untying them (model 3).
The performance of model 3 was much worse than  the complete model (model 1) as shown in Table \ref{ablation},
which demonstrates the need of sharing the weights for calculating matching scores  at both instance and class levels.

\subsubsection{Inconsistency Measurement}
As introduced in Section 4.5, ${\rm J}_{incon}$ is designed to measure the inconsistency among the representations of all support instances in a class.
After removing ${\rm J}_{incon}$, model 2 was optimized only using the objective function ${\rm J}_{match}$.
We can see that it performed much worse than the complete model. 
Furthermore, we calculated the mean of the Euclidean distances between every support instance pair  $(\widehat{\mathbf{s}}^i_k, \widehat{\mathbf{s}}^i_{k'})$ in the same class using model 1 and model 2 respectively.
For each support set, the calculation can be written as
\begin{equation}
	D = \frac{2}{N K (K-1)} \sum_{i=1}^{N}\sum_{k=1}^{K}\sum_{k'=k+1}^{K}|| \widehat{\mathbf{s}}^i_k - \widehat{\mathbf{s}}^i_{k'} ||_2^2.
\end{equation}
We sampled 20,000 support sets under the 5-way 5-shot configuration and calculated the mean of them.
The results were $0.0199$ and $0.0346$ for model 1 and model 2 respectively, which means that ${\rm J}_{incon}$ was effective at forcing the representations of the support instances in the same class to be close with each other.

${\rm J}_{incon}$ was further removed from model 5 and model 6 was obtained.
It can be found that the accuracy degradation from model 5 to model 6 was larger than the one from model 1 to model 2.
This implies that the ${\rm J}_{incon}$ objective function also benefited from the attentive aggregation over support instances.

\subsubsection{Local Matching}
First, the concatenation operation in local matching was removed from model 6 in this ablation study.
That's to say, instead of concatenating the representations of all support instances $\{\mathbf{S}_k\}_{k=1}^{K}$ into one single matrix as Eq. (\ref{concat}),
local matching was conducted between the query instance  and each support instance separately  to get their vector representations $\{(\widehat{\mathbf{s}}_k, \widehat{\mathbf{q}}_k); k=1,...,K\}$ (model 7).
It should be noticed that this led to $K$ different representations of a query instance according to each support class.
Then, the mean over $k$ for $\widehat{\mathbf{s}}_k$ and $\widehat{\mathbf{q}}_k$ were calculated to get the representations of the support set $\widehat{\mathbf{s}}$ and the query instance $\widehat{\mathbf{q}}$.
Comparing model 6 and model 7, we can see that the concatenation operation plays an important role in our model. 
One possible reason is that the concatenation operation can help local matching to restrain the  support instances with low similarity to the query.

\begin{figure}[!t]
	\centering
	\includegraphics[width=2.8in]{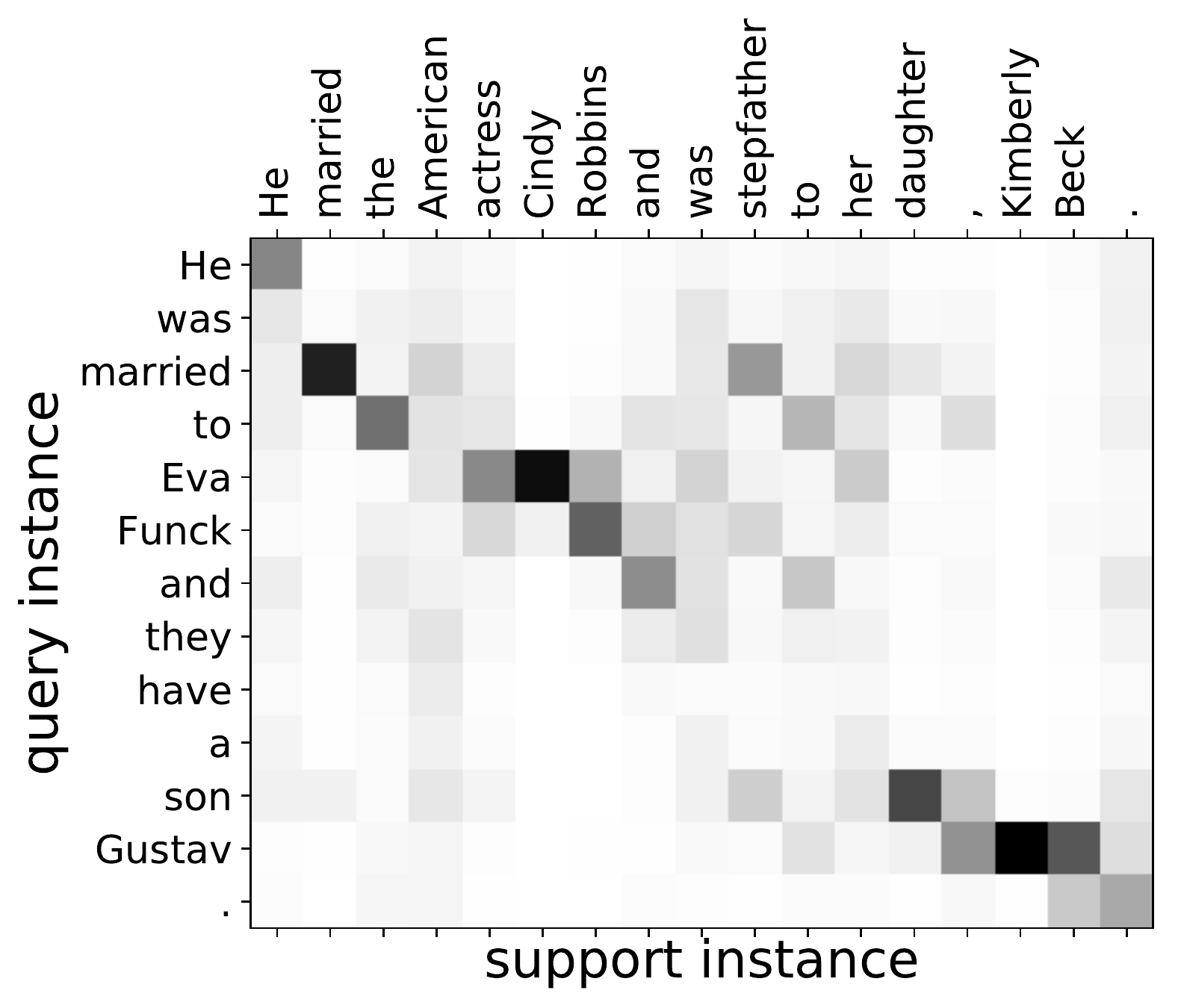}
	\caption{
The attention weight matrix calculated between the query instance and the support instance \#2 of class A in Table \ref{example}. The darker units have larger value.
The summation of one column in the matrix is one.
	}
	\label{attention}
\end{figure}

Second, the whole local matching module together with the concatenation and attentive aggregation operation were removed from model 6, which led to model 9.
Model 9 is similar to the one proposed by \citet{snell2017prototypical} that encoded the support and query instances independently.
The difference was that model 9 was equipped with more components, including  an LSTM layer, two pooling operations, and a learnable class matching function.
Comparing the performance of model 6 and model 9 in Table \ref{ablation}, we can see that the local matching operation significantly improves the performance in few-shot relation classification.
Fig. \ref{attention} shows the attention weight matrix calculated between the query instance and the support instance \#2 of class A in Table \ref{example}.
From this figure, we can see that the attention-based local matching is able to capture some matching relations of local contexts, such as the head entities \emph{Eva Funck} and \emph{Cindy Robbins}, the tail entities \emph{Gustav} and \emph{Kimberly Beck}, the key phrases \emph{son} and \emph{daughter}, the same keyword ``\emph{married}", and so on.

\subsubsection{Class Matching}
In this experiment, we compared two class matching functions, (1) Euclidean distance (ED) \cite{snell2017prototypical} and (2) a learnable MLP function as shown by Eq. (13).
In order to ignore the influence of the instance-level attentive aggregation, these two matching functions were compared based on model 6 and model 9.
After converting the MLP function in model 6 and model 9 to Euclidean distance, model 8 and model 10 were obtained.
Comparing the performance of these models in Table \ref{ablation}, we have two findings.
(1) When local matching was adopted, the learnable MLP for class matching (model 6) outperformed the ED metric (model 8) by a large margin.
(2) After removing local matching, the learnable MLP for class matching (model 9) performed not as good as  the ED metric (model 10).
One possible reason is that the local matching process enhances the interaction between a query instance and a support set when calculating $\widehat{\mathbf{s}}$ and $\widehat{\mathbf{q}}$.
Thus, simple Euclidean distance between them may not be able to describe the complex correlation and dependency between them.
On the other hand, MLP mapping is more powerful than calculating Euclidean distance, and can be more appropriate for class matching when local matching is also adopted.

\section{Conclusions}
In this paper, a neural network with multi-level matching and aggregation has been proposed  for few-shot relation classification.
First, the query and support instances are encoded interactively via local matching and aggregation.
Then, the support instances in a class are further aggregated to form the class prototype and the weights are calculated by attention-based instance matching.
Finally, a learnable MLP matching function is employed to calculate the class matching score between the query instance and each candidate class.
Furthermore, an additional objective function is designed to improve the consistency among the vector representations of all support instances in a class.
Experiments have demonstrated the effectiveness of our proposed model, which achieves state-of-the-art performance on the FewRel dataset.
Studying few-shot relation classification with data generated by distant supervision and extending our MLMAN model to zero-shot learning will be the tasks of our future work.

\section*{Acknowledgments}
We thank the anonymous reviewers for their valuable comments.
This work was partially funded by the National Nature Science Foundation of China (Grant No. U1636201, 61871358).

\bibliography{acl2019}

\begin{thebibliography}{33}
\expandafter\ifx\csname natexlab\endcsname\relax\def\natexlab#1{#1}\fi

\bibitem[{Bethard and Martin(2007)}]{Bethard2007Temporal}
Steven Bethard and James~H. Martin. 2007.
\newblock \href {http://aclweb.org/anthology/S07-1025} {Cu-tmp: Temporal
  relation classification using syntactic and semantic features}.
\newblock In \emph{Proceedings of the Fourth International Workshop on Semantic
  Evaluations (SemEval-2007)}, pages 129--132. Association for Computational
  Linguistics.

\bibitem[{Bowman et~al.(2015)Bowman, Angeli, Potts, and Manning}]{Bowman2015A}
Samuel~R. Bowman, Gabor Angeli, Christopher Potts, and Christopher~D. Manning.
  2015.
\newblock \href {https://doi.org/10.18653/v1/D15-1075} {A large annotated
  corpus for learning natural language inference}.
\newblock In \emph{Proceedings of the 2015 Conference on Empirical Methods in
  Natural Language Processing}, pages 632--642. Association for Computational
  Linguistics.

\bibitem[{Chen et~al.(2018)Chen, Ling, and Zhu}]{Chen2018Enhancing}
Qian Chen, Zhen-Hua Ling, and Xiaodan Zhu. 2018.
\newblock \href {http://aclweb.org/anthology/C18-1154} {Enhancing sentence
  embedding with generalized pooling}.
\newblock In \emph{Proceedings of the 27th International Conference on
  Computational Linguistics}, pages 1815--1826. Association for Computational
  Linguistics.

\bibitem[{Chen et~al.(2017)Chen, Zhu, Ling, Wei, Jiang, and
  Inkpen}]{Chen2017Enhanced}
Qian Chen, Xiaodan Zhu, Zhen-Hua Ling, Si~Wei, Hui Jiang, and Diana Inkpen.
  2017.
\newblock \href {https://doi.org/10.18653/v1/P17-1152} {Enhanced lstm for
  natural language inference}.
\newblock In \emph{Proceedings of the 55th Annual Meeting of the Association
  for Computational Linguistics (Volume 1: Long Papers)}, pages 1657--1668.
  Association for Computational Linguistics.

\bibitem[{Conneau et~al.(2017)Conneau, Kiela, Schwenk, Barrault, and
  Bordes}]{Conneau2017Supervised}
Alexis Conneau, Douwe Kiela, Holger Schwenk, Lo{\"i}c Barrault, and Antoine
  Bordes. 2017.
\newblock \href {https://doi.org/10.18653/v1/D17-1070} {Supervised learning of
  universal sentence representations from natural language inference data}.
\newblock In \emph{Proceedings of the 2017 Conference on Empirical Methods in
  Natural Language Processing}, pages 670--680. Association for Computational
  Linguistics.

\bibitem[{Finn et~al.(2017)Finn, Abbeel, and Levine}]{finn2017model}
Chelsea Finn, Pieter Abbeel, and Sergey Levine. 2017.
\newblock Model-agnostic meta-learning for fast adaptation of deep networks.
\newblock \emph{arXiv preprint arXiv:1703.03400}.

\bibitem[{Gao et~al.(2019)Gao, Han, Liu, and Sun}]{gao2019hybrid}
Tianyu Gao, Xu~Han, Zhiyuan Liu, and Maosong Sun. 2019.
\newblock Hybrid attention-based prototypical networks for noisy few-shot
  relation classification.

\bibitem[{Garcia and Bruna(2017)}]{garcia2017few}
Victor Garcia and Joan Bruna. 2017.
\newblock Few-shot learning with graph neural networks.
\newblock \emph{arXiv preprint arXiv:1711.04043}.

\bibitem[{Gong et~al.(2018)Gong, Qiu, Chen, Liang, and
  Huang}]{Gong2018Convolutional}
Jingjing Gong, Xipeng Qiu, Xinchi Chen, Dong Liang, and Xuanjing Huang. 2018.
\newblock \href {http://aclweb.org/anthology/D18-1186} {Convolutional
  interaction network for natural language inference}.
\newblock In \emph{Proceedings of the 2018 Conference on Empirical Methods in
  Natural Language Processing}, pages 1576--1585. Association for Computational
  Linguistics.

\bibitem[{Gong et~al.(2017)Gong, Luo, and Zhang}]{gong2017natural}
Yichen Gong, Heng Luo, and Jian Zhang. 2017.
\newblock Natural language inference over interaction space.
\newblock \emph{arXiv preprint arXiv:1709.04348}.

\bibitem[{Han et~al.(2018)Han, Zhu, Yu, Wang, Yao, Liu, and
  Sun}]{Han2018FewRel}
Xu~Han, Hao Zhu, Pengfei Yu, Ziyun Wang, Yuan Yao, Zhiyuan Liu, and Maosong
  Sun. 2018.
\newblock \href {http://aclweb.org/anthology/D18-1514} {Fewrel: A large-scale
  supervised few-shot relation classification dataset with state-of-the-art
  evaluation}.
\newblock In \emph{Proceedings of the 2018 Conference on Empirical Methods in
  Natural Language Processing}, pages 4803--4809. Association for Computational
  Linguistics.

\bibitem[{Hinton et~al.(2012)Hinton, Srivastava, Krizhevsky, Sutskever, and
  Salakhutdinov}]{hinton2012improving}
Geoffrey~E Hinton, Nitish Srivastava, Alex Krizhevsky, Ilya Sutskever, and
  Ruslan~R Salakhutdinov. 2012.
\newblock Improving neural networks by preventing co-adaptation of feature
  detectors.
\newblock \emph{arXiv preprint arXiv:1207.0580}.

\bibitem[{Hochreiter and Schmidhuber(1997)}]{Hochreiter1997Long}
S~Hochreiter and J~Schmidhuber. 1997.
\newblock Long short-term memory.
\newblock \emph{Neural Computation}, 9(8):1735--1780.

\bibitem[{Kim et~al.(2018)Kim, Hong, Kang, and Kwak}]{kim2018semantic}
Seonhoon Kim, Jin-Hyuk Hong, Inho Kang, and Nojun Kwak. 2018.
\newblock Semantic sentence matching with densely-connected recurrent and
  co-attentive information.
\newblock \emph{arXiv preprint arXiv:1805.11360}.

\bibitem[{Kim(2014)}]{Kim2014Convolutional}
Yoon Kim. 2014.
\newblock Convolutional neural networks for sentence classification.
\newblock \emph{Eprint Arxiv}.

\bibitem[{Koch et~al.(2015)Koch, Zemel, and Salakhutdinov}]{koch2015siamese}
Gregory Koch, Richard Zemel, and Ruslan Salakhutdinov. 2015.
\newblock Siamese neural networks for one-shot image recognition.
\newblock In \emph{ICML Deep Learning Workshop}, volume~2.

\bibitem[{Lowe et~al.(2015)Lowe, Pow, Serban, and Pineau}]{Lowe2015The}
Ryan Lowe, Nissan Pow, Iulian Serban, and Joelle Pineau. 2015.
\newblock \href {https://doi.org/10.18653/v1/W15-4640} {The ubuntu dialogue
  corpus: A large dataset for research in unstructured multi-turn dialogue
  systems}.
\newblock In \emph{Proceedings of the 16th Annual Meeting of the Special
  Interest Group on Discourse and Dialogue}, pages 285--294. Association for
  Computational Linguistics.

\bibitem[{Mintz et~al.(2009)Mintz, Bills, Snow, and
  Jurafsky}]{mintz2009distant}
Mike Mintz, Steven Bills, Rion Snow, and Dan Jurafsky. 2009.
\newblock Distant supervision for relation extraction without labeled data.
\newblock In \emph{Proceedings of the Joint Conference of the 47th Annual
  Meeting of the ACL and the 4th International Joint Conference on Natural
  Language Processing of the AFNLP: Volume 2-Volume 2}, pages 1003--1011.
  Association for Computational Linguistics.

\bibitem[{Mishra et~al.(2018)Mishra, Rohaninejad, Chen, and
  Abbeel}]{Mishra2018A}
Nikhil Mishra, Mostafa Rohaninejad, Xi~Chen, and Pieter Abbeel. 2018.
\newblock A simple neural attentive meta-learner.

\bibitem[{Mueller and Thyagarajan(2016)}]{mueller2016siamese}
Jonas Mueller and Aditya Thyagarajan. 2016.
\newblock Siamese recurrent architectures for learning sentence similarity.
\newblock In \emph{AAAI}, volume~16, pages 2786--2792.

\bibitem[{Munkhdalai and Yu(2017)}]{munkhdalai2017meta}
Tsendsuren Munkhdalai and Hong Yu. 2017.
\newblock Meta networks.
\newblock \emph{arXiv preprint arXiv:1703.00837}.

\bibitem[{Nair and Hinton(2010)}]{Nair2010Rectified}
Vinod Nair and Geoffrey~E. Hinton. 2010.
\newblock Rectified linear units improve restricted boltzmann machines.
\newblock In \emph{International Conference on International Conference on
  Machine Learning}.

\bibitem[{Pennington et~al.(2014)Pennington, Socher, and
  Manning}]{Pennington2014Glove}
Jeffrey Pennington, Richard Socher, and Christopher Manning. 2014.
\newblock \href {https://doi.org/10.3115/v1/D14-1162} {Glove: Global vectors
  for word representation}.
\newblock In \emph{Proceedings of the 2014 Conference on Empirical Methods in
  Natural Language Processing (EMNLP)}, pages 1532--1543. Association for
  Computational Linguistics.

\bibitem[{Ravi and Larochelle(2016)}]{ravi2016optimization}
Sachin Ravi and Hugo Larochelle. 2016.
\newblock Optimization as a model for few-shot learning.

\bibitem[{Santoro et~al.(2016)Santoro, Bartunov, Botvinick, Wierstra, and
  Lillicrap}]{santoro2016meta}
Adam Santoro, Sergey Bartunov, Matthew Botvinick, Daan Wierstra, and Timothy
  Lillicrap. 2016.
\newblock Meta-learning with memory-augmented neural networks.
\newblock In \emph{International conference on machine learning}, pages
  1842--1850.

\bibitem[{Snell et~al.(2017)Snell, Swersky, and Zemel}]{snell2017prototypical}
Jake Snell, Kevin Swersky, and Richard Zemel. 2017.
\newblock Prototypical networks for few-shot learning.
\newblock In \emph{Advances in Neural Information Processing Systems}, pages
  4077--4087.

\bibitem[{Vinyals et~al.(2016)Vinyals, Blundell, Lillicrap, Wierstra
  et~al.}]{vinyals2016matching}
Oriol Vinyals, Charles Blundell, Timothy Lillicrap, Daan Wierstra, et~al. 2016.
\newblock Matching networks for one shot learning.
\newblock In \emph{Advances in neural information processing systems}, pages
  3630--3638.

\bibitem[{Wang et~al.(2016)Wang, Cao, de~Melo, and Liu}]{Wang2016Relation}
Linlin Wang, Zhu Cao, Gerard de~Melo, and Zhiyuan Liu. 2016.
\newblock \href {https://doi.org/10.18653/v1/P16-1123} {Relation classification
  via multi-level attention cnns}.
\newblock In \emph{Proceedings of the 54th Annual Meeting of the Association
  for Computational Linguistics (Volume 1: Long Papers)}, pages 1298--1307.
  Association for Computational Linguistics.

\bibitem[{Xiong et~al.(2018)Xiong, Yu, Chang, Guo, and Wang}]{Xiong2018One}
Wenhan Xiong, Mo~Yu, Shiyu Chang, Xiaoxiao Guo, and William~Yang Wang. 2018.
\newblock \href {http://aclweb.org/anthology/D18-1223} {One-shot relational
  learning for knowledge graphs}.
\newblock In \emph{Proceedings of the 2018 Conference on Empirical Methods in
  Natural Language Processing}, pages 1980--1990. Association for Computational
  Linguistics.

\bibitem[{Zelenko et~al.(2002)Zelenko, Aone, and
  Richardella}]{Zelenko2002Kernel}
Dmitry Zelenko, Chinatsu Aone, and Anthony Richardella. 2002.
\newblock \href {http://aclweb.org/anthology/W02-1010} {Kernel methods for
  relation extraction}.
\newblock In \emph{Proceedings of the 2002 Conference on Empirical Methods in
  Natural Language Processing (EMNLP 2002)}.

\bibitem[{Zeng et~al.(2014)Zeng, Liu, Lai, Zhou, and Zhao}]{Zeng2014Relation}
Daojian Zeng, Kang Liu, Siwei Lai, Guangyou Zhou, and Jun Zhao. 2014.
\newblock \href {http://aclweb.org/anthology/C14-1220} {Relation classification
  via convolutional deep neural network}.
\newblock In \emph{Proceedings of COLING 2014, the 25th International
  Conference on Computational Linguistics: Technical Papers}, pages 2335--2344.
  Dublin City University and Association for Computational Linguistics.

\bibitem[{Zhou et~al.(2016)Zhou, Shi, Tian, Qi, Li, Hao, and
  Xu}]{Zhou2016Attention}
Peng Zhou, Wei Shi, Jun Tian, Zhenyu Qi, Bingchen Li, Hongwei Hao, and Bo~Xu.
  2016.
\newblock \href {https://doi.org/10.18653/v1/P16-2034} {Attention-based
  bidirectional long short-term memory networks for relation classification}.
\newblock In \emph{Proceedings of the 54th Annual Meeting of the Association
  for Computational Linguistics (Volume 2: Short Papers)}, pages 207--212.
  Association for Computational Linguistics.

\bibitem[{Zhou et~al.(2018)Zhou, Li, Dong, Liu, Chen, Zhao, Yu, and
  Wu}]{Zhou2018Multi}
Xiangyang Zhou, Lu~Li, Daxiang Dong, Yi~Liu, Ying Chen, Wayne~Xin Zhao, Dianhai
  Yu, and Hua Wu. 2018.
\newblock \href {http://aclweb.org/anthology/P18-1103} {Multi-turn response
  selection for chatbots with deep attention matching network}.
\newblock In \emph{Proceedings of the 56th Annual Meeting of the Association
  for Computational Linguistics (Volume 1: Long Papers)}, pages 1118--1127.
  Association for Computational Linguistics.

\end{thebibliography}
\bibliographystyle{acl_natbib}
\end{document}